\documentclass[letterpaper,journal]{IEEEtran}
\usepackage{amsmath,amsfonts}
\usepackage{algorithmic}
\usepackage{algorithm}
\usepackage{array}
\usepackage[caption=false,font=normalsize,labelfont=sf,textfont=sf]{subfig}
\usepackage{textcomp}
\usepackage{stfloats}
\usepackage{url}
\usepackage[hidelinks]{hyperref}
\usepackage{verbatim}
\usepackage{graphicx}
\usepackage{cite}
\usepackage{xcolor}
\usepackage{soul}
\usepackage{multirow}
\usepackage{tabularx}
\usepackage{gensymb}
\begin{document}

\title{Yummy Operations Robot Initiative: An Autonomous Cooking System Utilizing a Modular Robotic Kitchen and Dual-Arm Proprioceptive Manipulator}

\author{Donghun Noh, Hyunwoo Nam, Kyle Gillespie, Yeting Liu, Dennis Hong}

\maketitle

\begin{abstract}
This paper presents Yummy Operations Robot Initiative (YORI), a proprioceptive dual-arm robotic system that demonstrates autonomous multi-dish cooking for scalable food service applications. YORI integrates a dual-arm manipulator equipped with proprioceptive actuators, custom-designed tools, appliances, and a structured kitchen environment to address the complexities of cooking tasks. The proprioceptive actuators enable fast, precise, force-controlled movements while mitigating the risks associated with cooking-related impacts. The system's modular kitchen design and flexible tool-changing mechanism support simultaneous multi-dish preparation through torque control and optimization-based motion planning and scheduling. A comprehensive scheduling framework with dynamic rescheduling ensures reliable adaptation to new orders and delays. The system was publicly validated through live demonstrations, reliably preparing steak-frites across multiple convention sessions. This paper details YORI's design and explores future directions in kitchen optimization, task planning, and food quality control, demonstrating its potential as a scalable robotic cooking solution. A system introduction and cooking videos are available online\footnote{%
\url{https://youtu.be/KIkVHRenZ8A}, 
\url{https://youtu.be/M_xZkmoz2WE}, 
\url{https://youtu.be/np8yNOKwJTk}, 
\url{https://youtu.be/80aalf3CG58}}.
\end{abstract}

\section{Introduction}
\IEEEPARstart{E}{fforts} to develop autonomous robotic cooking systems are driven by the fast-paced demands of contemporary life. In a world where time is often scarce, and labor shortages and aging populations are becoming pressing societal issues, these robots represent both technological efficiency and a milestone toward maintaining and improving quality of life. These robots assist individuals with limited cooking skills or physical challenges, offering independence and access to diverse culinary experiences. Beyond personal use, these robots have the potential to revolutionize the culinary industry, enabling restaurants and food services to enhance efficiency, maintain consistent quality, and explore new recipes. Moreover, they align with societal goals such as reducing food waste and improving cooking energy efficiency.

For these reasons, developing a robotic cooking system has been actively explored in the robotics field since the 1980s, progressing from single-task machines like Suzumo's sushi robots to increasingly sophisticated systems. Recent commercial developments include specialized robots like Flippy by Miso Robotics for burger flipping, while research efforts have produced more versatile systems such as Samsung's Bot Chef and Moley's Robotic Kitchen. However, existing systems either remain single-task specialists or, in the case of the multi-task systems, are still in developmental phases without demonstrating fully autonomous multi-dish cooking capabilities. Table~\ref{tab:cooking_robots_comparison} compares the key characteristics of existing robotic cooking systems.

\begin{table}[t!]
\centering
\scriptsize{}
\caption{Comparison of Autonomous Cooking Robot Systems}
\label{tab:cooking_robots_comparison}
\begin{tabular}{l|c|c|c|c|c}
\hline
\textbf{System} & \textbf{Tasks$^a$} & \textbf{Arm/Base$^b$} & \textbf{Gripper$^c$} & \textbf{Kitchen$^d$} & \textbf{Apps.$^e$} \\
\hline
\hline
Moley & M & D/C & A & H & 2-3 \\
Bot Chef & M & D/C & 3F & H & 1-2 \\
Flippy & S & S/F & T & S & 1-2 \\
\textbf{YORI} & \textbf{M} & \textbf{D/R} & \textbf{T} & \textbf{MS} & \textbf{10+} \\
\hline
\noalign{\vskip 2pt}
\multicolumn{6}{l}{\footnotesize $^a$M: Multiple, S: Single}\\
\multicolumn{6}{l}{\footnotesize $^b$D: Dual, S: Single, C: Ceiling, F: Fixed, R: Rotating}\\
\multicolumn{6}{l}{\footnotesize $^c$A: Anthropomorphic, 3F: 3-Finger, T: Tool changer}\\
\multicolumn{6}{l}{\footnotesize $^d$H: Human, S: Structured, MS: Modular Structured}\\
\multicolumn{6}{l}{\footnotesize $^e$Number of custom appliances}
\end{tabular}
\vspace{-6mm}
\end{table}

Today's cooking robots are actively trying to utilize advanced tactile sensors, force/torque sensors, and camera sensors such as RGB-D cameras or LiDAR sensors, as well as increasingly leveraging AI in order to enhance personalization and adaptability. The use of collaborative robots (cobots) is also an effort to create flexible cooking robot systems that can interact accurately and safely around humans in dynamic environments. However, there is still a long way to go before fully replicating the intricacies and complexities seen throughout cooking operations.

\begin{figure*}[t!]
\centering
\includegraphics[width=1\textwidth]{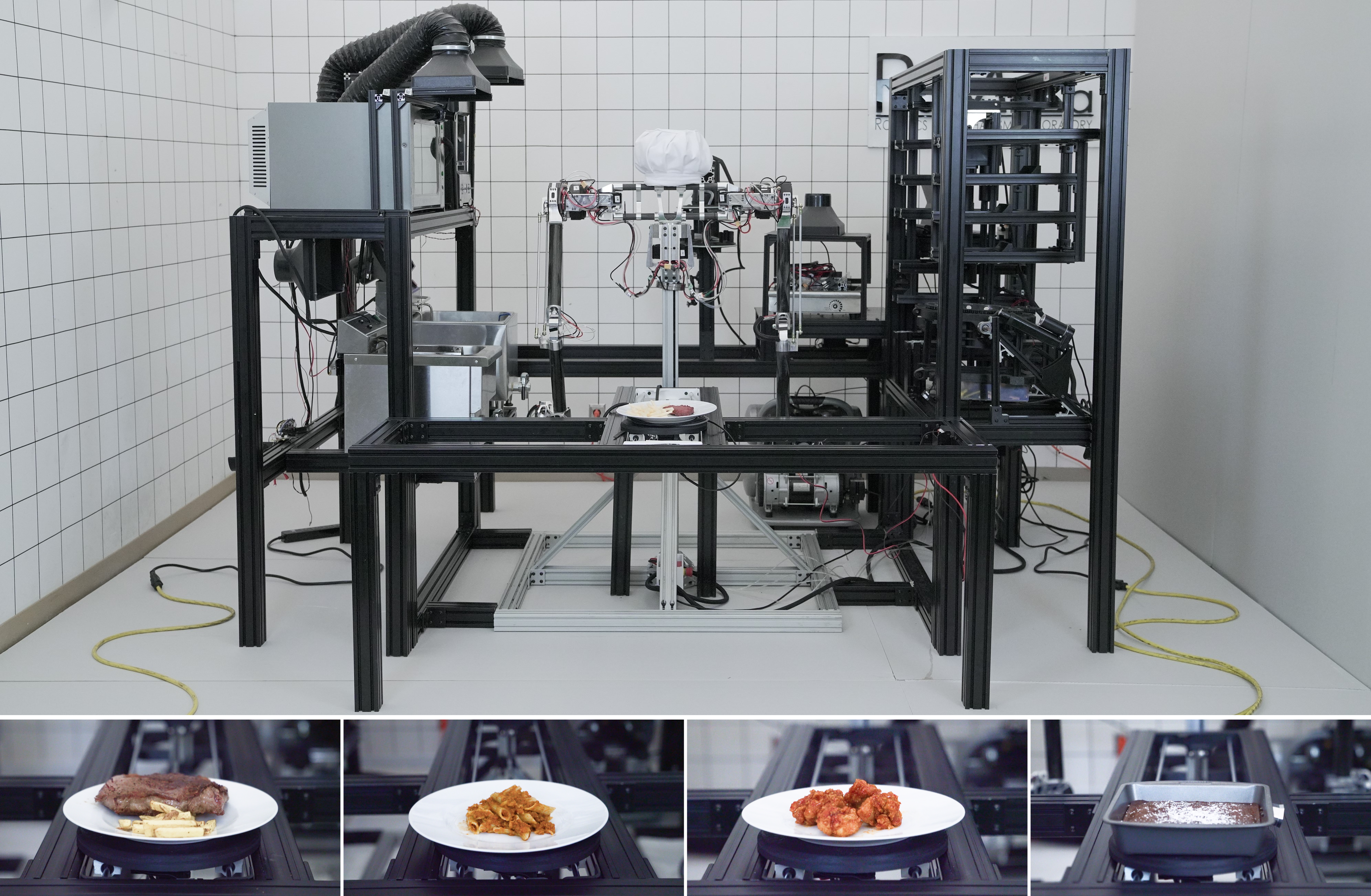}
\caption{Overview of the YORI System, featuring a dual-arm manipulator equipped with proprioceptive actuators at its center. This setup is encircled by four modular kitchen units, each outfitted with customized or newly developed tools and appliances, positioned at the front, back, and sides. Examples of dishes cooked by the system are shown below, from left to right: Steak Frites, Tomato Penne Pasta, Spicy Fried Chicken, and Brownies.}
\label{fig:yori_system}
\vspace{-4mm}
\end{figure*}

In this article, we present our autonomous robotic cooking system, YORI (Yummy Operations Robot Initiative), devised to tackle these challenges through a fusion of cutting-edge robotics technologies and an innovative approach aimed at simplifying cooking tasks. The overview of the system is shown in Fig.~\ref{fig:yori_system}. This approach focuses on leveraging the unique characteristics of robots, diverging from conventional human-centric cooking methods. Through this method, we designed kitchen tools and appliances suitable for robotic use and created a kitchen layout that maximizes the workspace of the system's dual-arm manipulator. Additionally, to prioritize the diversity and scalability of the cooking menu, we developed a modular kitchen design that allows for not only general kitchen setups but also specialized  configurations tailored to specific dishes. Finally, we enhanced the system's autonomy, stability, and versatility by developing a dual-arm manipulator equipped with proprioceptive actuators. These actuators are fast and precise backdrivable force-sensing motors that allow for the compliant manipulation essential for cooking motions.

This article contributes to the field by presenting a comprehensive exploration of the design and implementation of an autonomous robotic cooking system. The main features of this article are as follows: Section~\ref{background} lays out the design and algorithms that were developed and implemented, in conjunction with existing research; Section~\ref{hardware} details the hardware design, encompassing the design of the dual-arm manipulator, kitchen tools and appliances, and the modular kitchen layout; Section~\ref{software} discusses the software running the system and its implementation; Sections~\ref{conclusion},~\ref{limitations}, and~\ref{future_work} present the system validation, current limitations, and future research directions, respectively.

\section{Background}\label{background}

\subsection{Dual-Arm Manipulator}
Cobots emerged in the late 1990s, enabling safe human-robot collaboration in tasks such as organizing, assembly and inspection. While lighter and more responsive than traditional industrial robots, cobots still suffer from relatively slow acceleration and high stiffness that limit performance in dynamic environments, and their weight remains inefficient for mobile platforms. These limitations have driven research toward lighter, backdrivable, and compliant manipulators with improved force control, exemplified by Berkeley's Quasi-Direct Drive (QDD) BLUE~\cite{gealy2019quasi} and KoreaTech's tendon-driven LIMS~\cite{kim2017anthropomorphic}.

\subsection{Proprioceptive Actuator}
To enable backdrivability and compliance in manipulators, proprioceptive actuators, also known as QDD, offer an efficient solution. These actuators combine high-torque brushless DC motors with low reduction gearboxes (typically <10:1) to maintain torque transparency~\cite{Proprioceptive1}. Unlike traditional high-geared servo motors that suffer from high reflected inertia and poor compliance, proprioceptive actuators excel at backdrivability and impact mitigation~\cite{Proprioceptive2}. Their torque transparency enables direct torque estimation from motor current, eliminating the need for external force/torque sensors and simplifying both hardware and control systems.

\subsection{Robotic Cell Layout Design}
Robotic cell layout design extends traditional manufacturing layout optimization principles while addressing unique constraints of robotic systems, particularly in multi-robot environments and human-robot interaction scenarios~\cite{zhang2017challenges}. Recent advances in computational power have enabled sophisticated optimization techniques including Mixed Integer Linear Programming, Genetic Algorithms, and Convex Optimization to solve complex workspace design problems, with simulation-based optimization and digital twin technologies allowing virtual validation before physical implementation.

\subsection{Manipulator Control}
Accurate modeling is crucial for controlling manipulators with proprioceptive actuators, enabling precise torque calculation and force feedback for stable operation in dynamic environments. This accurate modeling enables two key control strategies: inverse dynamics control for trajectory tracking~\cite{lynch2017modern} and impedance control for safe interaction tasks~\cite{impedance}

Inverse dynamics control in the joint space is as follows:
\begin{equation}
\tau_{d} = M(q) \ddot{q} + C(q, \dot{q}) + g(q) + J^T F_{tip}
\end{equation}
\noindent where $M(q)$ is the mass matrix, $C(q, \dot{q})$ is the term containing Coriolis and centripetal effects, $g(q)$ is the gravitational terms, $J$ is the Jacobian, and $F_{tip}$ is the end-effector contact forces. The friction term was neglected for simplicity.

Impedance control in Cartesian space is shown in the equation below:
{\normalsize
\begin{multline}
\tau_d = M(q) \ddot{q} + C(q, \dot{q}) + g(q)\\ + J^T(K(x_d - x) + D(\dot{x}_d - \dot{x}) + F_{tip})
\end{multline}
}
\noindent where $K$, $D$ are stiffness and damping matrices, and $x$, $\dot{x}$ are the position and velocity in Cartesian space.

\subsection{Motion Planning}
\label{background:motion_planning}
Minimizing jerk (time derivative of acceleration) is essential for smooth trajectory generation. Flash and Hogan showed human arm movements naturally minimize jerk using $L_2$ norm optimization~\cite{hogan1985}, while Yazdani demonstrated that $L_\infty$ norm minimization yields biologically plausible "bang-bang" control policies~\cite{cvxopt_min_jerk}. Although sampling-based methods with time parameterization are computationally efficient in confined environments, they often sacrifice smoothness and energy efficiency, leading to continued research in optimization-based techniques with jerk constraints~\cite{10610437}. Our work leverages YORI's structured kitchen environment to eliminate dynamic obstacles, enabling full exploitation of minimum jerk optimization for smooth, energy-efficient dual-arm trajectories.

\subsection{Task Scheduling}
Effective task scheduling and real-time adaptability are crucial for robust automated cooking systems. The kitchen can be modeled as a finite state machine of heterogeneous robotic appliances, with orders represented as finite sequences containing operational parameters like cooking duration and seasoning amounts. These constraints, formulated using linear temporal logic (LTL) and computation tree logic (CTL), naturally align with job-shop scheduling problems (JSSPs). Prior work has applied dispatching rules~\cite{KitchenScheduling} and online MIP formulations~\cite{yi2022online} to cooking JSSPs. Building on these foundations, our work incorporates dynamic rescheduling logic and adaptive programming strategies to better handle real-world complexities including delays, failures, and continuous order arrivals.

\section{Hardware Design}\label{hardware}
During the early stages of the project, we developed a prototype dual-arm manipulator using servo actuators to verify the feasibility of performing cooking tasks using robotic manipulators~\cite{YORI1}. 
However, due to the usage of rigid servo actuators and lack of force/torque sensors, the robot manipulator was unable to complete certain tasks where a precise torque estimation or rapid movement was required, such as cutting or tossing food. Furthermore, since all the components including actuators and structural parts are off-the-shelf products without any design optimizations, the payload at the end-effector was limited and the overall workspace was not ideal. 
With the numerous constructive experiences from the servo motor actuated manipulator prototype, we developed the next-generation proprioceptive dual-arm manipulator using high-performance quasi-direct drive actuators. These actuators allow for the manipulators compatibility with many cooking tasks, in particular those requiring impacts and precise force control, while being compliant to changes in the constructed environment.

\subsection{Dual-Arm Manipulator} 
An initial servo-actuated prototype validated the joint configuration and link dimensions for cooking tasks~\cite{YORI1}. Building on this verified kinematic design, a next-generation dual-arm manipulator was developed using proprioceptive actuators, enabling impact-resilient manipulation with precise force control and environmental compliance, which are capabilities essential for dynamic cooking operations such as cutting and tossing.

\subsubsection{Hardware Configuration}
The YORI dual-arm manipulator features 11 Degrees of Freedom (DoF) total: one torso yaw joint and two 5-DoF arms. Each arm comprises shoulder pitch/yaw, elbow pitch, and wrist pitch/roll joints. The 5-DoF configuration was deliberately chosen over traditional 6-7 DoF designs to minimize arm mass, enhancing dynamic performance and payload capacity for cooking tasks. A pneumatic tool changer at each wrist enables rapid switching between cooking implements.

Link dimensions were determined based on human upper body proportions~\cite{human_measure} for compatibility with existing kitchen tools: upper arm (0.400 m), forearm (0.375 m), and clavicle (0.420 m) lengths allow both coordinated dual-arm tasks and independent operation. Carbon fiber tubes were selected for all arm links due to their optimal strength-to-weight ratio. The 1.3 m platform height ensures full workspace coverage. Table~\ref{tab:parameters} summarizes the mechanical parameters.

\begin{table}[t]
\centering
\caption{Dual-Arm Manipulator Mechanical Parameters}
\begin{tabular}{l c|l c}
\hline
\textbf{Link} & \textbf{Mass [kg]} & \textbf{Link} & \textbf{Length [m]} \\
\hline
\hline
Clavicle & 1.457 & Clavicle & 0.420 \\
Shoulder & 2.154 & Shoulder & 0.150 \\
Upper arm & 1.570 & Upper arm & 0.400 \\
Forearm & 0.854 & Forearm & 0.375 \\
Wrist & 0.733 & Wrist & 0.085 \\
\hline
\end{tabular}
\label{tab:parameters}
\end{table}

\subsubsection{Actuator Selection}
Joint torque requirements were determined through static moment analysis for worst-case loading conditions (fully extended horizontal configuration)~\cite{YORI1}. Based on cooking task analysis, a 3 kg payload capacity was specified with a 1.5 safety factor, yielding maximum torque requirements of 37.93 Nm (shoulder pitch), 20.29 Nm (elbow pitch), and 3.75 Nm (wrist pitch). Three proprioceptive actuator types from Westwood Robotics were selected to meet these specifications, as shown in Table~\ref{tab:actuators}.

Dynamic simulations with actual cooking trajectories were conducted to validate the actuator selection. The system was verified to handle 5 kg payloads, 67\% above the design requirement, while maintaining torques well within continuous ratings, as shown in Table~\ref{tab:max_torque}. This operational headroom ensures reliable performance across diverse cooking scenarios.

\begin{table}[t]
\centering
\caption{Proprioceptive Actuator Specifications}
\begin{tabular}{l c c c}
\hline
\textbf{Parameter} & \textbf{Type 1} & \textbf{Type 2} & \textbf{Type 3}\\
\hline
\hline
Mass [kg] & 0.650 & 0.925 & 0.250 \\
Gear ratio & 6:1 & 6:1 & 9:1 \\
Continuous torque [Nm] & 16.8 & 33.0 & 4.2 \\
Peak torque [Nm] & 33.5 & 67.0 & 10.5 \\
Speed constant [RPM/V] & 27.3 & 7.1 & 14.3 \\
\hline
\multicolumn{4}{l}{Type 1: Torso yaw, Shoulder yaw}\\
\multicolumn{4}{l}{Type 2: Shoulder pitch, Elbow pitch}\\
\multicolumn{4}{l}{Type 3: Wrist pitch, Wrist roll}\\
\hline
\end{tabular}
\label{tab:actuators}
\end{table}

\begin{table}[t]
\centering
\scriptsize{}
\caption{Torque Verification with 5 kg Payload}
\begin{tabular}{l c c}
\hline
\textbf{Joint} & \textbf{Avg. torque [Nm]} & \textbf{Max. torque [Nm]} \\
\hline
\hline
Torso yaw & 1.49 & 12.22 \\
Shoulder yaw & 0.82 & 6.65 \\
Shoulder pitch & 13.30 & 31.32 \\
Elbow pitch & 15.63 & 23.97 \\
Wrist pitch & 3.50 & 4.59 \\
Wrist roll & 0.91 & 2.20 \\
\hline
\end{tabular}
\label{tab:max_torque}
\end{table}

\begin{figure}[t]
\centering
\includegraphics[width=\columnwidth]{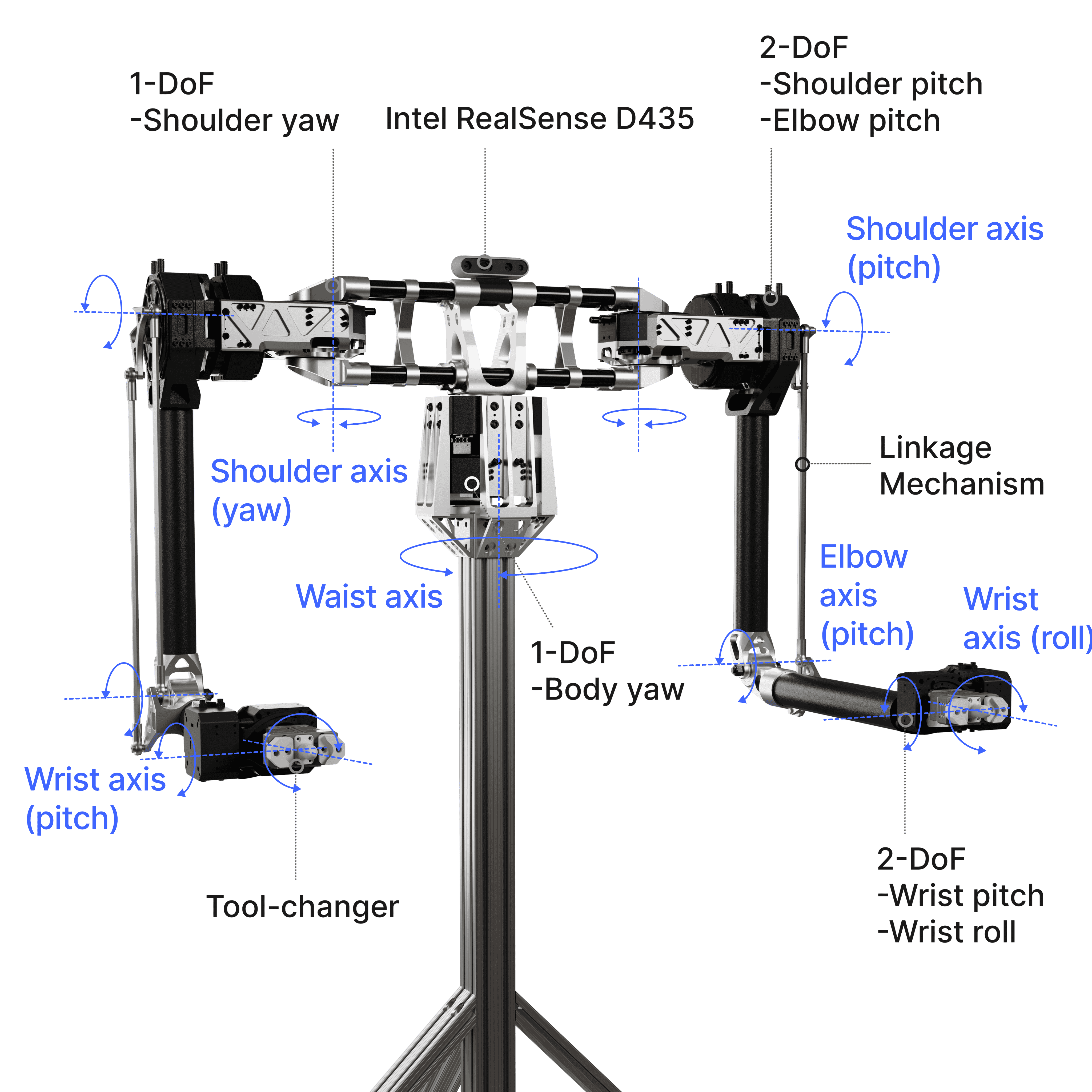}
\caption{Dual-arm manipulator with proprioceptive actuators and a linkage mechanism, designed for dynamic and dexterous manipulation with low arm inertia.}
\label{fig:manipulator}
\end{figure}

\begin{table}[t]
\centering
\caption{Joint Range of Motion}
\begin{tabular}{l c | l c}
\hline
\textbf{Joint} & \textbf{Range [°]} & \textbf{Joint} & \textbf{Range [°]} \\
\hline
\hline
Torso yaw & ±180 & Elbow pitch & -150/+95 \\
Shoulder yaw & ±130 & Wrist pitch & -133/+111 \\
Shoulder pitch & ±180 & Wrist roll & ±180 \\
\hline
\end{tabular}
\label{tab:range_of_motion}
\end{table}

\subsubsection{Joint and Linkage}
To minimize inertia for fast and accurate control, the elbow pitch actuator was relocated near the shoulder joint. The torso yaw, shoulder yaw, and shoulder pitch actuators remain directly mounted at their respective joints due to their proximity to the torso. The wrist pitch and roll actuators were kept at the joint to maintain design clarity and reduce system complexity, as their relatively small size and weight have minimal impact on arm dynamics.

Since the elbow pitch actuator was relocated, a mechanism was required to transmit torque to the elbow joint. While timing belts offer simple power transmission, their low stiffness introduces compliance that limits torque control bandwidth~\cite{katz}. Therefore, a dual 4-bar linkage mechanism in parallelogram configuration was implemented as shown in Fig.~\ref{fig:manipulator}.

This design transmits torque from the relocated actuator to the elbow joint with a 1:1 ratio while maintaining high stiffness. The double linkage configuration maximizes the elbow's range of motion (-150° to +95°) and eliminates singularities at full extension, critical for reaching across the kitchen workspace. The full range of motion is presented in Table~\ref{tab:range_of_motion}.

\begin{figure}[t]
\centering
\includegraphics[width=0.9\columnwidth]{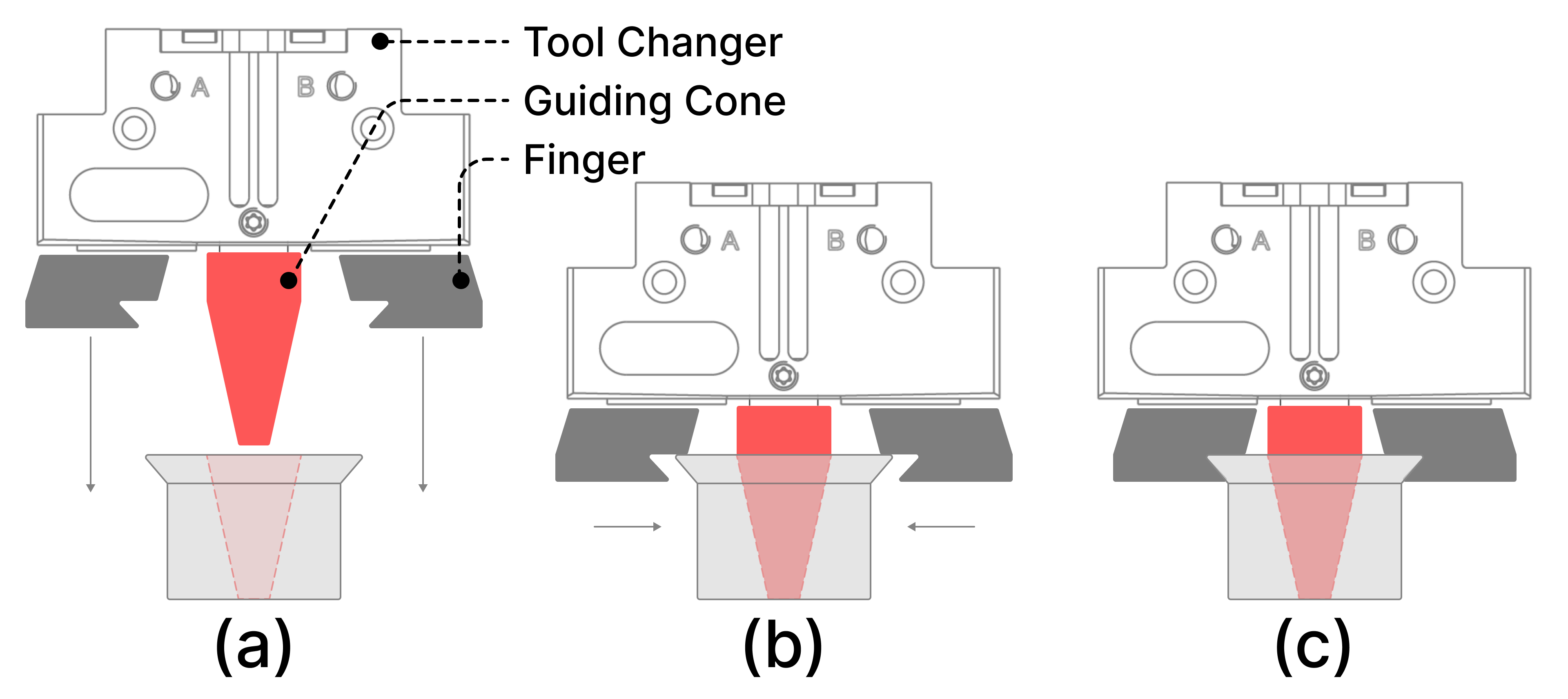}
\caption{Tool Changer Mechanism: (a) A guiding cone aligns the tool changer to correct grasping position, (b) Tool changer's fingers are actuated by using compressed air to grip, (c) Fully locked tool changer and tool plate}
\label{fig:toolchanger}
\end{figure}

\subsubsection{Tool Changing System}
A manipulator with a general-purpose end-effector could be designed to handle several tasks, but would eventually limit the scope of the system. However, a manipulator that can utilize multiple specialized end-effectors is capable of being more effective in handling any number of tasks~\cite{toolchanger}. To allow YORI to better handle various tools for different tasks, YORI utilizes a pneumatic tool changer system. This allowed YORI to swap end-effectors between various cooking tools including a pan, a container, a knife, and a meat tenderizer. Each tool employs a custom tool plate, which the tool changer's fingers can grip and lock onto using compressed air as shown in Fig.~\ref{fig:toolchanger}. A guiding cone feature allows the end-effector to correct any offset between the correct grasping position and the current position before the finger surface touches the tool plate. From our experiments, the tool changer could successfully guide itself to the locking position on the tool plate with up to 8 mm of translational offset from the center and with up to 10 degrees of angular offset.

\subsubsection{Camera}
The camera attached on the torso is an Intel RealSense D435. The set RGB frame resolution was 640 x 480 at 30 frame rate, and the RGB sensor resolution is 2-megapixel. The Intel RealSense D435 is also an active stereo depth camera with an ideal range of 0.3 m to 3 m, which can be utilized for future research.

\subsection{Modular Kitchen} 
\begin{figure}[t]
\centering
\includegraphics[width=0.8\columnwidth]{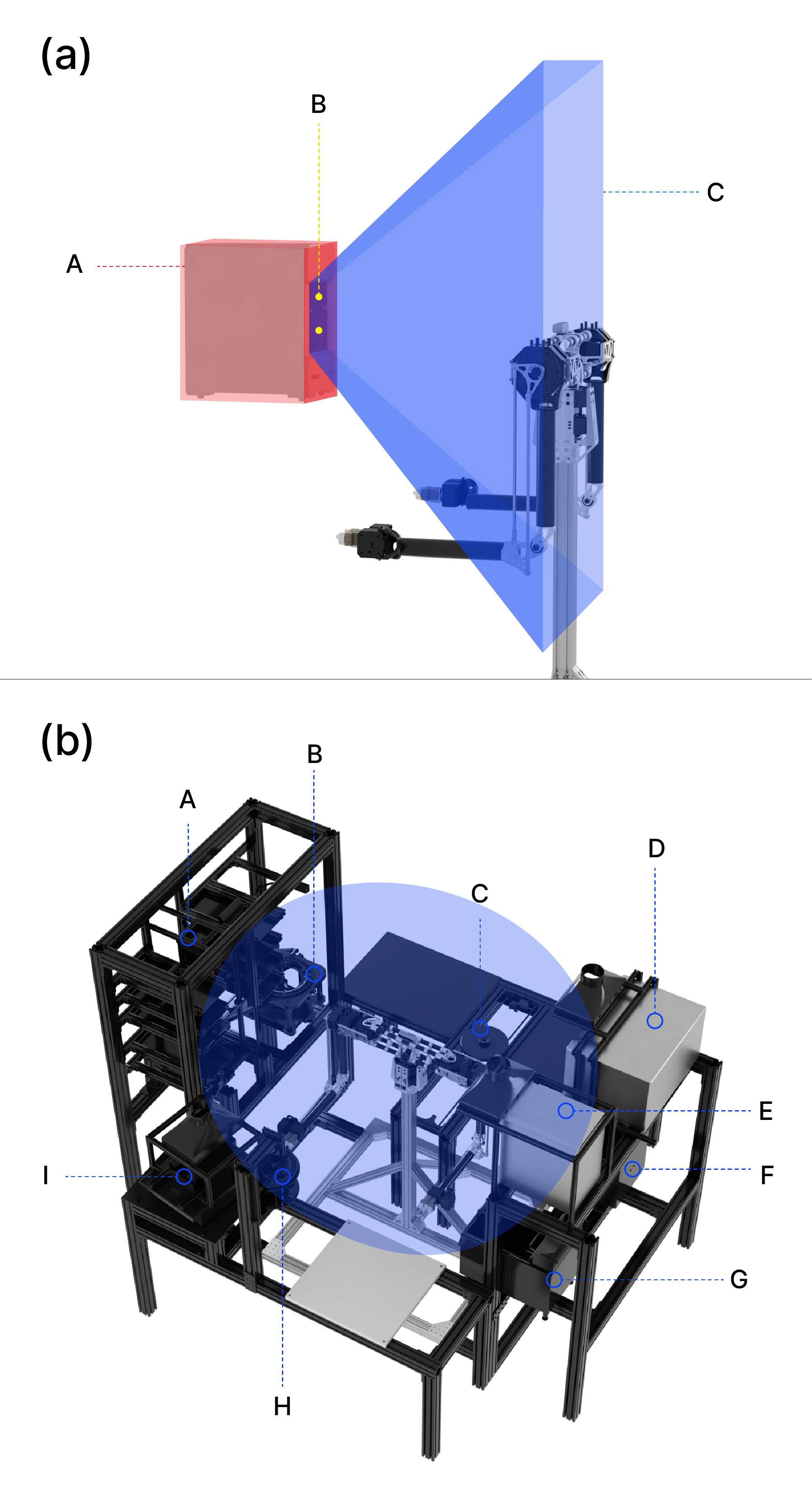}
\caption{Layout design of the cooking cell. (a) Layout components. A: red cube representing appliance size and position, B: yellow key points for end-effector targets, C: blue cuboid defining arm movement space between appliance and manipulator. (b) Final cooking cell configuration with manipulator workspace (blue ellipsoid) and appliances. A: storage shelf, B: rotating pot, C: dish trolley, D: convection oven, E: salamander broiler, F: deep fryer, G: pasta cooker, H: spice dispenser, I: induction cooktop.}
\label{fig:layout}
\end{figure}

The modular kitchen design enables organized, scalable cooking operations through optimized layout planning. The kitchen layout was formulated as an optimization problem to maximize workspace efficiency. As illustrated in Fig.~\ref{fig:layout}(a), each appliance is represented by a red cube (A), with yellow dots (B) marking key points and blue cuboids (C) defining arm movement spaces.

\begin{equation}
\begin{aligned}
\min_{\mathbf{x}} \ & \mathbf{J}=   \sum_{i}  ||\mathbf{x_i} - \mathbf{v}||_2\\
\textrm{s.t.} \quad & (\mathbf{x}_i - \mathbf{v})^T \mathbf{A} (\mathbf{x}_i - \mathbf{v}) \leq 1, \, \forall i \\
\quad & f_{\text{OBB}}(R_i, R_j) = 0, \, \forall i \neq j \\
\quad & f_{\text{OBB}}(R_i, B_j) = 0, \, \forall i, j \\
\end{aligned}
\end{equation}

where $\mathbf{x}_i = [x_i, y_i, z_i]^T$ represents the coordinates of the key point for each kitchen appliance. $\mathbf{v} = [x_C, y_C, z_C]^T$ denotes the center of the manipulator's workspace. $\mathbf{A}$ is a $3 \times 3$ symmetric positive definite matrix defining the shape and orientation of the ellipsoid that represents the manipulator's workspace. $R_i$ and $B_j$ are the $i^{th}$ red cubic and the $j^{th}$ blue cuboid, respectively, representing different appliances or components within the kitchen environment. $f_{\text{OBB}}$ is a function that uses the Oriented Bounding Box (OBB) algorithm to determine the overlap status between cuboids, returning 0 if there is no overlap.

The objective function minimizes the total distance from each appliance's key point to the workspace center, optimizing for efficiency and accessibility. The ellipsoidal workspace constraint ensures all operation points remain within the manipulator's reach, while non-overlap constraints implemented through $f_{\text{OBB}}$ prevent physical interference between appliances. The optimization results were refined with heuristic adjustments for manufacturing feasibility, hygiene requirements, and workflow efficiency, producing the final layout shown in Fig.~\ref{fig:layout}(b).

\subsection{Custom Tools and Appliances}
Full integration and networking of cooking appliances was essential for system optimization. Rather than having the manipulator directly operate analog controls, which would introduce multiple failure points, an Internet of Things (IoT) approach was adopted~\cite{IOT_Review}. Appliance selection was guided by four criteria: safety, robotic operability, cleanability, and multifunctionality. Based on common cooking operations, appliances were either modified from commercial units or custom-designed, as detailed in Table~\ref{tab:appliances}.

\begin{table}[t]
\caption{Appliance Selection}
\begin{center}
\begin{tabular}{c c c}
\hline
\textbf{Crucial Operations}& \textbf{Appliance} & \textbf{Source}\\
\hline\hline
Baking & Convection Oven & Modify\\
Searing & Salamander Broiler & Modify\\
Grilling & Induction Cooktop & Modify\\
Boiling/Steaming & Pasta Cooker & Modify\\
Frying & Deep Fryer & Modify\\
Cutting & Food Processor & Design\\
Pureeing & Food Processor & Design\\
Stir Frying & Rotating Mixer & Design\\
Mixing & Rotating Mixer & Design\\
Seasoning & Spice Dispenser & Design\\
\hline
\end{tabular}
\label{tab:appliances}
\end{center}
\end{table}

\begin{figure*}[t]
\centering
\includegraphics[width=\textwidth]{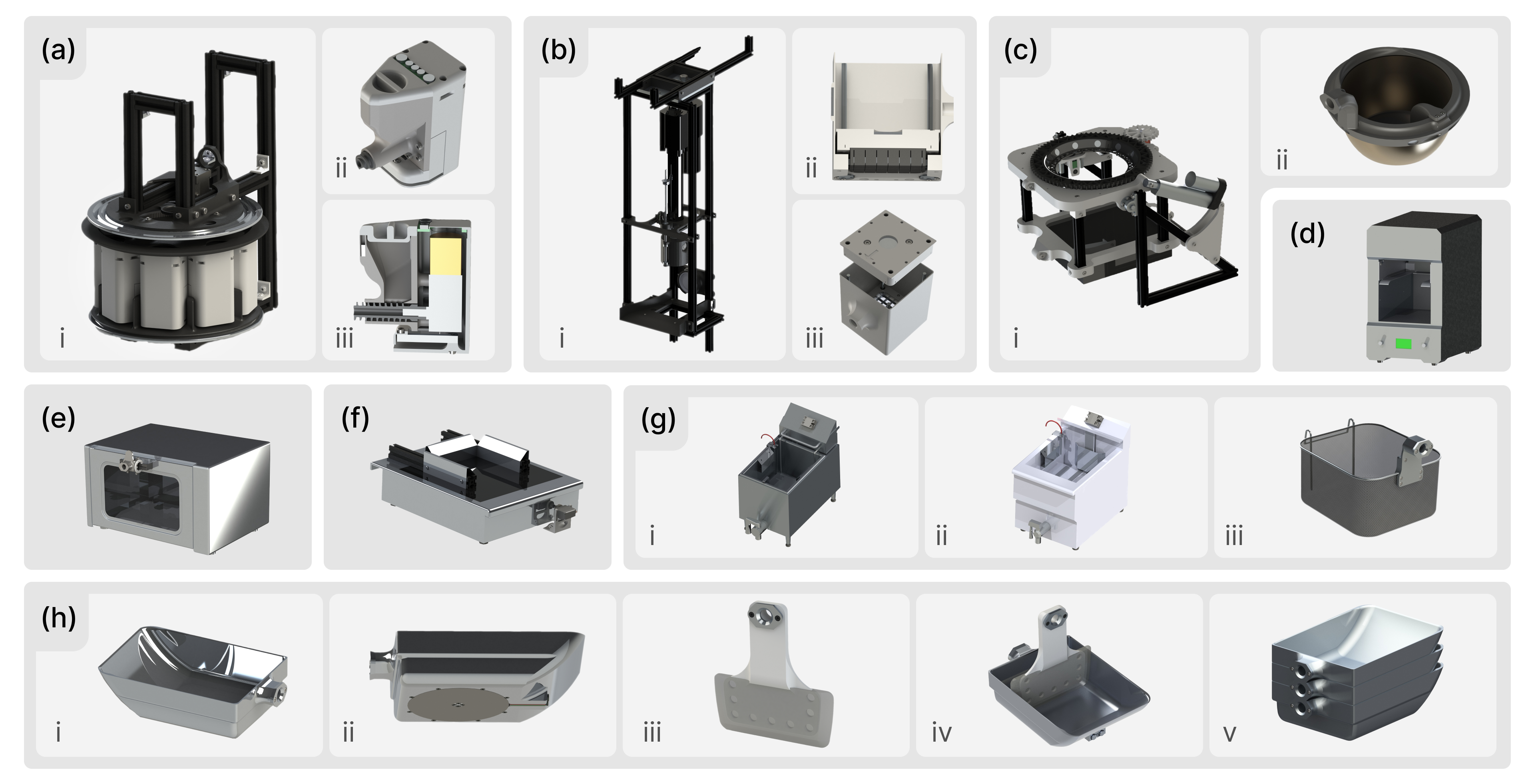}
\caption{YORI kitchen appliances and specialized tools: (a-c) Preparation appliances/tools: spice dispenser with modules, food processor with dicing chamber, rotating mixer with detachable pot. (d-g) Cooking appliances/tools: salamander broiler, convection oven, induction cooktop, deep fryer/water boiler with basket. (h) Custom induction pan with squeegee tool.}
\label{fig:Appliances}
\end{figure*}

Commercial appliances with high heat or electrical requirements were modified for robotic control to leverage their proven reliability. The convection oven, salamander broiler, induction cooktop, pasta cooker, and deep fryer shown in Fig.~\ref{fig:Appliances}(d-g) were retrofitted by replacing analog controls with digital counterparts, as detailed in Table~\ref{tab:digital_control}.

\begin{table}[t]
\caption{Appliance Retrofitting Techniques}
\begin{center}
\begin{tabular}{c c c}
\hline
\textbf{Original Component}
&\textbf{Digital Counterpart}
&\textbf{Relevant Appliance}\\
\hline\hline
Thermostat
&\begin{tabular}{@{}c@{}}Solid State Relay \\ Thermocouple \end{tabular}
&\begin{tabular}{@{}c@{}c@{}}Convection Oven \\ Deep Fryer \\ Pasta Cooker \end{tabular}\\
\hline
Potentiometer Knob
&\begin{tabular}{@{}c@{}}Potentiometer IC \\ (e.g., MCP4018T)\end{tabular} 
& Induction Cooktop\\
\hline
Encoder Knobs
&\begin{tabular}{@{}c@{}}Paired optocoupler ICs \\ (e.g., VO2601)\end{tabular}
& Salamander Broiler\\
\hline
\end{tabular}
\label{tab:digital_control}
\end{center}
\end{table}

\subsubsection{Rotating Mixer}
The rotating mixer addresses mixing and tumbling operations that would otherwise limit multitasking capabilities if performed directly by the manipulator~\cite{Stir_Fry_arms}. As shown in Fig.~\ref{fig:Appliances}(c), the design features two rotation axes for varied movement patterns: fixed angles for stir-frying and swinging motions for mixing. The modular drum system enables quick tool changes and simplified cleaning.

The drum's primary axis rotation is driven through magnetic coupling between the pot and mixer interfaces, ensuring repeatable insertion/removal while preventing slippage. A magnetic encoder provides speed feedback and alignment control, while an IR thermal camera enables contactless temperature monitoring during operation.

\subsubsection{Food Processor}
The food processor employs a modular design addressing safety concerns associated with exposed blades~\cite{Robot_Cutting}. As illustrated in Fig.~\ref{fig:Appliances}(b), removable dicing chambers and blade assemblies facilitate cleaning and functional flexibility. During operation, pre-loaded ingredients in the dicing chamber are extruded through blades via a linear actuator-driven lid, with precise guidance from constraining rails. An electromagnet locks the chamber in place, while the manipulator positions a collection vessel below. This modularity supports various blade configurations for dicing, slicing, and pureeing operations.

\subsubsection{Spice Dispenser}
The spice dispenser provides precise, repeatable dispensing of powdered ingredients through eight modular units mounted in a rotating hub, as shown in Fig.~\ref{fig:Appliances}(a). Each module contains an Archimedes screw driven by a geared DC motor, with a load cell measuring weight changes for accurate dosing. Neodymium magnets with metallic coating provide both power transmission and I2C communication between modules and the central hub. The 360-degree hub rotation enables even distribution over targets and facilitates automated module exchange from a single access point.

\subsubsection{Pan and Squeegee}
A custom induction-compatible pan was designed for efficient ingredient transfer, as illustrated in Fig.~\ref{fig:Appliances}(h). The pan features a curved ramp for squeegee-assisted transfer, stackable geometry with internal slopes and external ledges, and a torus-shaped stainless steel disk addressing thermal expansion mismatches with the aluminum body. Temperature monitoring is achieved through a thermocouple connected via Alnico magnets chosen for high-temperature resistance, enabling magnetic docking for electrical connection to the measurement system.

\section{Software \& Implementation} \label{software}
\begin{figure*}[t!]
\centering
\includegraphics[width=0.8\textwidth]{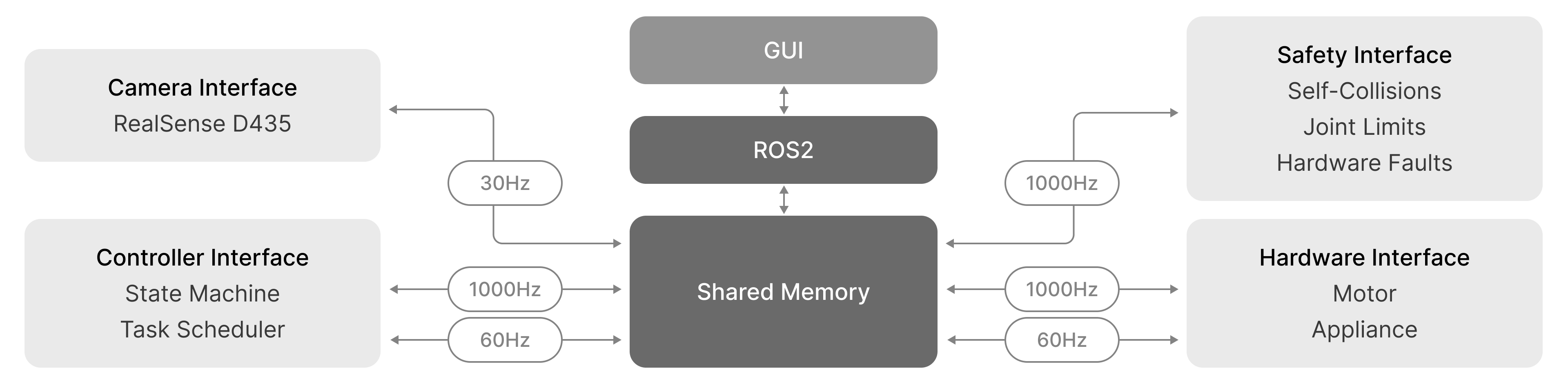}
\caption{Overview of the YORI System's software architecture, highlighting interfaces critical for high-frequency operations like safety checks, state checks, and motor control, communicating at 1000Hz. Conversely, components not requiring high-frequency interactions, such as camera systems, task scheduler, and appliance control, function at 30Hz or 60Hz. To guarantee swift communication and enhance system stability, data processing is executed via shared memory. Remote commands for the system are accepted through WiFi, leveraging the ROS2 middleware, solely upon specific order reception.}
\label{fig:yori_software_system}
\end{figure*}

The YORI system's software architecture implements a hierarchical framework with multiple processes that orchestrate dual-arm manipulation, appliance control, and task scheduling. As illustrated in Fig.~\ref{fig:yori_software_system}, the system comprises four primary interfaces: Camera Interface, Controller Interface, Safety Interface, and Hardware Interface. Critical operations including safety monitoring and motor control execute at 1000 Hz for precise manipulator control, while camera processing, task scheduling, and appliance control operate at 30-60 Hz. Inter-process communication leverages shared memory for data exchange between processes, with synchronization mechanisms ensuring data consistency during concurrent access.

\subsection{Manipulator Control}

\begin{figure*}[t!]
\centering
\includegraphics[width=0.8\textwidth]{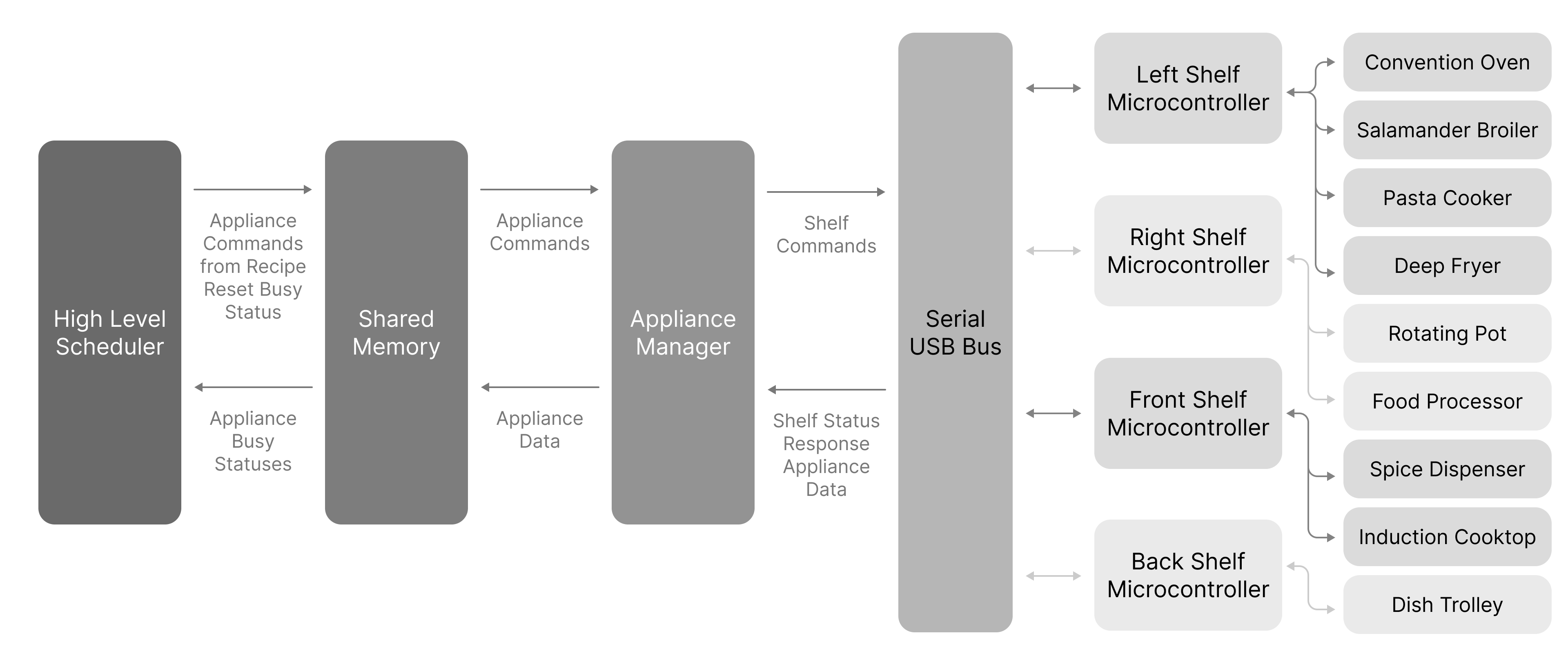}
\caption{Flow diagram illustrating the control structure and messages used to control the integrated appliances in the kitchen cell. The High Level Scheduler communicates commands and statuses with the appliance manager through the systems shared memory. The appliance manager in turn communicates with each local microcontroller one at a time over the shared usb bus by sending commands tied to one of the unique microcontroller IDs.}
\label{fig:Appliance_manager}
\end{figure*}

\subsubsection{Control Method}
Within the YORI system, cooking actions are primarily divided into two categories: transporting motions for ingredient and tool movement utilize inverse dynamics control, while grasping motions for tool latching and environmental interactions employ impedance control. This dual-mode approach enables seamless transitions between rigid control during aerial transportation and compliant behavior during contact operations, accommodating environmental uncertainties while preventing excessive contact forces.

\subsubsection{Payload Estimation}
Proprioceptive actuators enable real-time external force estimation by comparing measured torques with expected values based on arm configuration and dynamics. This capability facilitates automatic payload detection upon grasping and serves as a fault detection mechanism. Collisions manifest as unexpected force increases, while failed grasps are indicated by absent expected forces, enabling immediate response to anomalous conditions.

\subsubsection{Dual-Arm Control}
The dual-arm configuration divides into two kinematic chains starting from either the torso or shoulder yaw joint depending on task requirements. Each arm maintains independent controllers with task-specific stiffness values. For instance, during pan sweeping, the pan-holding arm maintains high stiffness for stable positioning while the squeegee-wielding arm operates compliantly to maintain consistent contact force with the pan surface. This heterogeneous control enables one arm to toss fries compliantly while the other transports ingredients with rigid control.

\subsection{Tool Changer Control}
The tool changer employs pneumatic actuation through directional control valves for binary open/close control of the gripper fingers. The control system operates at 100 Hz synchronized with manipulator movements, ensuring reliable tool grasping and release during cooking operations.

\subsection{Robotic Kitchen Control} 
The YORI kitchen system accommodates its modularity through a three-level appliance management system, visualized in Fig.~\ref{fig:Appliance_manager}. At the lowest level, microcontrollers with unique IDs directly control individual appliances through a shared USB connection. The system employs standardized commands (\textit{Read}, \textit{Command}, \textit{Initialize}) to request appliance status, set new goals, and establish serial connections respectively. Each appliance maintains unique input commands and status outputs, with the universal \textit{Busy} status flag serving as the primary synchronization mechanism. The code controlling each appliance is modular and non-blocking, enabling multiple appliances to operate in parallel while the controller maintains continuous serial bus access.

The intermediate appliance manager maintains USB connections and coordinates data transfer through a shared memory database divided into command and status sections for each appliance. An \textit{Update} variable indicates scheduler modifications: when changed from 0 to 1, the manager transmits new commands to the appropriate microcontroller, then resets the variable to 0, signaling readiness for new scheduler writes. This interchange prevents simultaneous write conflicts while maintaining system responsiveness. The manager queries appliance status at regular intervals, writing responses to corresponding memory allocations.

At the highest level, the scheduler parses recipe files as sequences of parameterized actions coupled with gate conditions. Each action targets a specific arm or appliance, with progression determined by gate conditions: timed delays, arm trajectory completions, or \textit{Busy} status changes. This abstraction enables recipe portability across different kitchen configurations while maintaining execution reliability through the shared memory architecture.

\subsection{Motion Planning} 
\begin{figure}[t]
\centering
\includegraphics[width=\columnwidth]{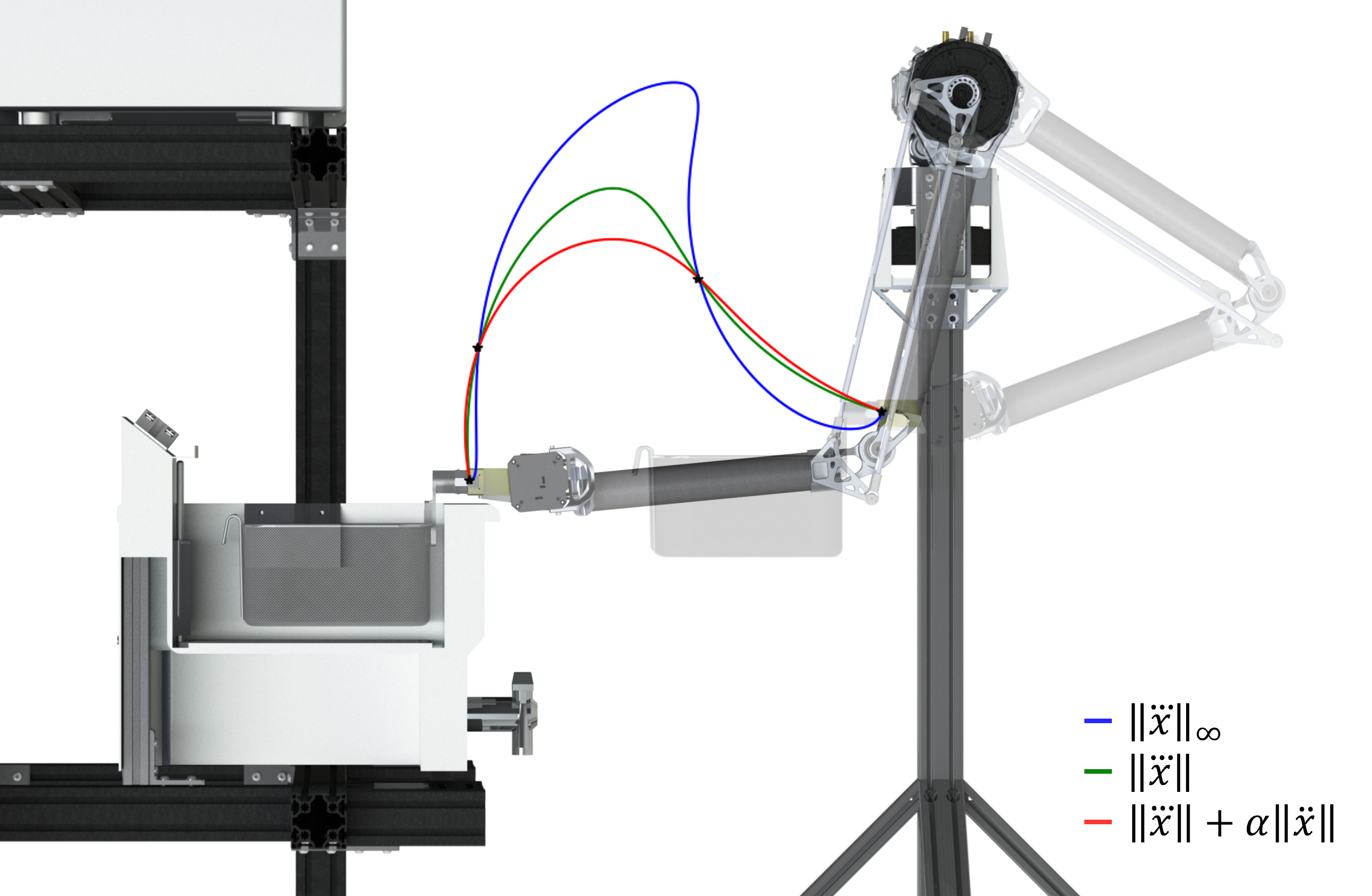}
\caption{Optimized trajectories with via points using different objectives for taking out a fryer basket from the deep fryer. The blue line represents the optimization objective with $L_\infty$ norm of jerk profile, the green line represents objective with $L_2$ norm of jerk profile, and the red line represents the objective with combined $L_2$ norms of jerk and weighted acceleration profile, $\alpha = 0.01$. Via points are represented as black stars.}
\label{fig:optimization}
\end{figure}

Initial bang-bang control implementations revealed trajectory discontinuities at via points, causing undesirable overshooting. The refined approach formulates trajectory generation as a quadratic programming problem:

\begin{equation}
\begin{aligned}
\min_{\mathbf{x}} \quad & ||\dddot{\mathbf{x}}||_2 + \alpha ||\ddot{\mathbf{x}}||_2\\
\textrm{s.t.} \quad & A\mathbf{x} = B\\
\end{aligned}
\end{equation}

\noindent where $\dddot{\mathbf{x}}$ represents jerk and $\ddot{\mathbf{x}}$ represents acceleration profiles, with constraints $A\mathbf{x} = B$ representing boundary conditions for position, velocity, and acceleration. Fig.~\ref{fig:optimization} demonstrates the practical impact through fryer basket extraction, where adding weighted acceleration ($\alpha = 0.01$) significantly improves the trajectory in terms of both smoothness and 
travel distance compared to pure minimum-jerk solutions.

\subsection{Task Scheduling}
The system models cooking operations as a job-shop scheduling problem using Google OR-Tools' CP-SAT solver~\cite{cpsatlp}. Beyond traditional JSSP constraints, the formulation incorporates cooking-specific requirements.

Concurrence constraints encode manipulator reachability limits. For example, the manipulator cannot simultaneously access the deep fryer on the left shelf and the dicer on the right shelf. Additionally, some operations require continuous tending, such as holding containers beneath the food processor during dicing. Horizon constraints ensure timely order completion and prevent dishes from cooling after oven removal.

Algorithm~\ref{CP-SAT} details these constraints transformed into linear, interval, and logical formulations. The CP-SAT solver employs parallel portfolio strategies including Lazy Clause Generation, SAT search, LP Branching, and Pseudo-Cost branching, with boolean variables assigning task order within schedules.

\setlength{\textfloatsep}{2pt}
\begin{algorithm}[t]
\caption{CP-SAT Approach to Cooking JSSP}
{\footnotesize
\begin{algorithmic}[1]
\STATE$x_{i,j,m} \gets \text{start time of task j, recipe i, run on machine m}$
\\$\text{where }i\in\{0,\dots,q\},j\in\{0,\dots,t\},m\in\{0,\dots,r\}$\\
\STATE$d_{i,j} \gets \text{duration of }x_{i,j}, ml_i \gets \text{order deadline}$
\STATE$O_{i,j} \gets \text{position of task j, recipe i from start to finish}$
\\$\text{where }O_{i,j}\in\{0,\dots,f | f=\sum_{i=0}^q t\} $
\STATE$e_{i,j}=x_{i,j,m}+d_{i,j}$
\\$\textbf{Define Constraints:}$
\STATE$\text{Precedence: } x_{i,j} \geq e_{i,j-1}$
\STATE$\text{Overlap: } x_{i,j,m} \leq x_{k,l,m}+d_{i,j,m}$
\STATE$\text{Concurrence: } e_{i,j,m} \leq x_{k,l,n}\;\text{Or}\;e_{k,l,n} \leq x_{i,j,m} : \{(m,n) \in \{\text{Incompatible Pairings}\}\}$
\STATE$\text{Horizon: }x_{i,0}+e_{i,t}\leq x_{i,0}+ml_i\;\&\ e_{i,t}<e_{i+1,t} $
\\$\textbf{Set Objective:}$
\STATE$\text{Minimize: }e_{q,t}$
\\$\textbf{Run Solver:}$
\FOR{\text{Set Number of Propagators or Timespan}}
    \STATE$\text{Establish domains: }[[O_{i,j}\leq [j,\dots,t]]]$
    \STATE$\text{Run Propagation of Variables Until Failure or Success}$
    \\$\text{ie. } O_{0,0}=0 \Rightarrow O_{1,0}=2\Rightarrow \dots \Rightarrow O_{0,t}=f$
    \\ $\text{returns failure} x_{0,0}+e_{0,t}>x_{0,0}+ml_0$
    \WHILE{Failure}
        \STATE$\text{Explanation for Failure: } O_{0,0}=0 \Rightarrow  O_{0,t}\neq f $
        \STATE$\text{Backtrack and Update Domains}$
        \STATE$\text{Propagate from Backtrack Until Failure or Success}$
    \ENDWHILE
    \STATE$\text{Store Successful Solution}$
\ENDFOR
\STATE$\text{Return Most Optimal Successful Solution}$
\end{algorithmic}
}
\label{CP-SAT}
\end{algorithm}

\begin{algorithm}[t!]
\caption{Heuristic Algorithm for Scheduling Strategy}
{\footnotesize
\begin{algorithmic}[1]
\WHILE{\text{YORI is Running}}
    \STATE $\text{Check for new recipe order inputs to add to to-do list}$ 
    \STATE $\text{Check for faults during task completion}$ 
    \STATE $\text{Check for completed tasks to remove from to-do list}$ 
    \IF{\text{A currently running task has been completed}}
        \STATE $\text{Remove task from currently running list and add to}$
        \STATE $\text{finished task list}$ 
        \STATE $\text{Add next task to run on the currently running tasks}$
        \STATE $\text{Remove said task from the to-do list}$
    \ENDIF
    \IF{\text{New order or fault}}
        \IF{\text{New fault}}
            \IF{\text{Task tries is less than max tries}}
                \STATE $\text{Add task back to to-do list}$
            \ELSE
                \STATE $\text{Remove all tasks related to jobs that still need}$
                \STATE $\text{the relevant machine from to-do list}$
            \ENDIF
        \ENDIF
    \STATE $\text{Rerun the CP-SAT solver with the updated task list}$        
    \STATE $\text{Adjust machines to-do list}$
    \ENDIF
\ENDWHILE
\end{algorithmic}
}
\label{Heuristic}
\end{algorithm}

\begin{figure*}[t!]
\centering
\includegraphics[width=1\textwidth]{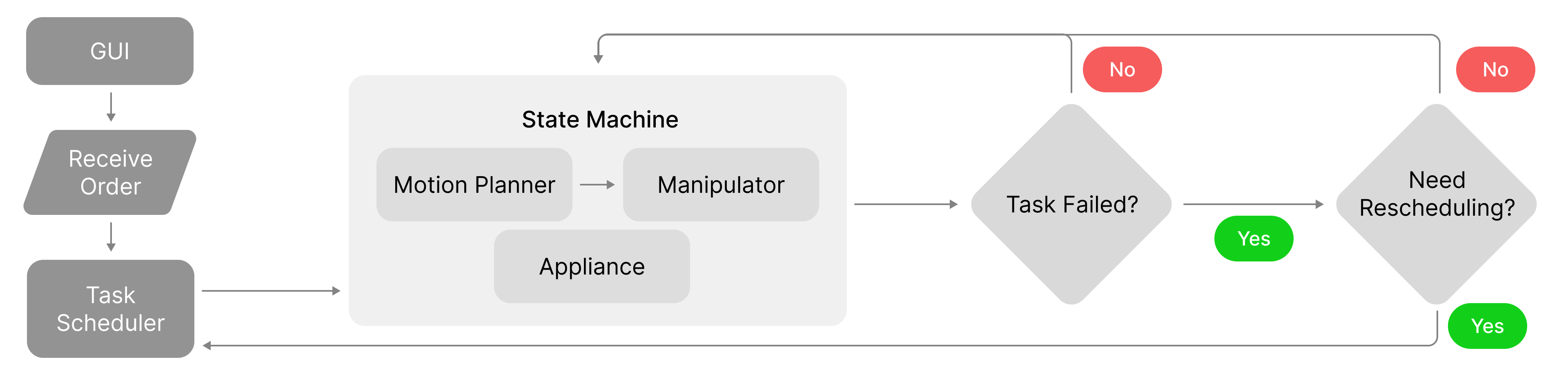}
\caption{An overview of the dynamic replanning sequence. The Task Scheduler will resolve for the optimal schedule under two conditions. The first being the input of a new recipe to the queue from YORI's associated GUI. The second condition is when the state machine governing the running of the manipulator and appliances detects that a task that it has been given by the scheduler has failed and can no longer be retried.}
\label{fig:planning}
\end{figure*}

\subsection{Dynamic Replanning}
Real-world operation demands adaptive scheduling for asynchronous events. The system implements dynamic replanning triggered by new order arrivals or task execution failures, as illustrated in Fig.~\ref{fig:planning} and detailed in Algorithm~\ref{Heuristic}.

\subsubsection{Fault Handling}
Each task encodes acceptable retry attempts based on risk assessment. Low-risk failures like tool grasping misalignment trigger realignment with decreased impedance gains. The scheduler readjusts timing for subsequent tasks upon each retry. Mechanical failures requiring unavailable machines result in affected recipe cancellation, with the system optimizing maintenance pause placement to minimize lost time.

\subsubsection{New Order Placed}
\begin{figure*}[t]
\centering
\includegraphics[width=1\textwidth]{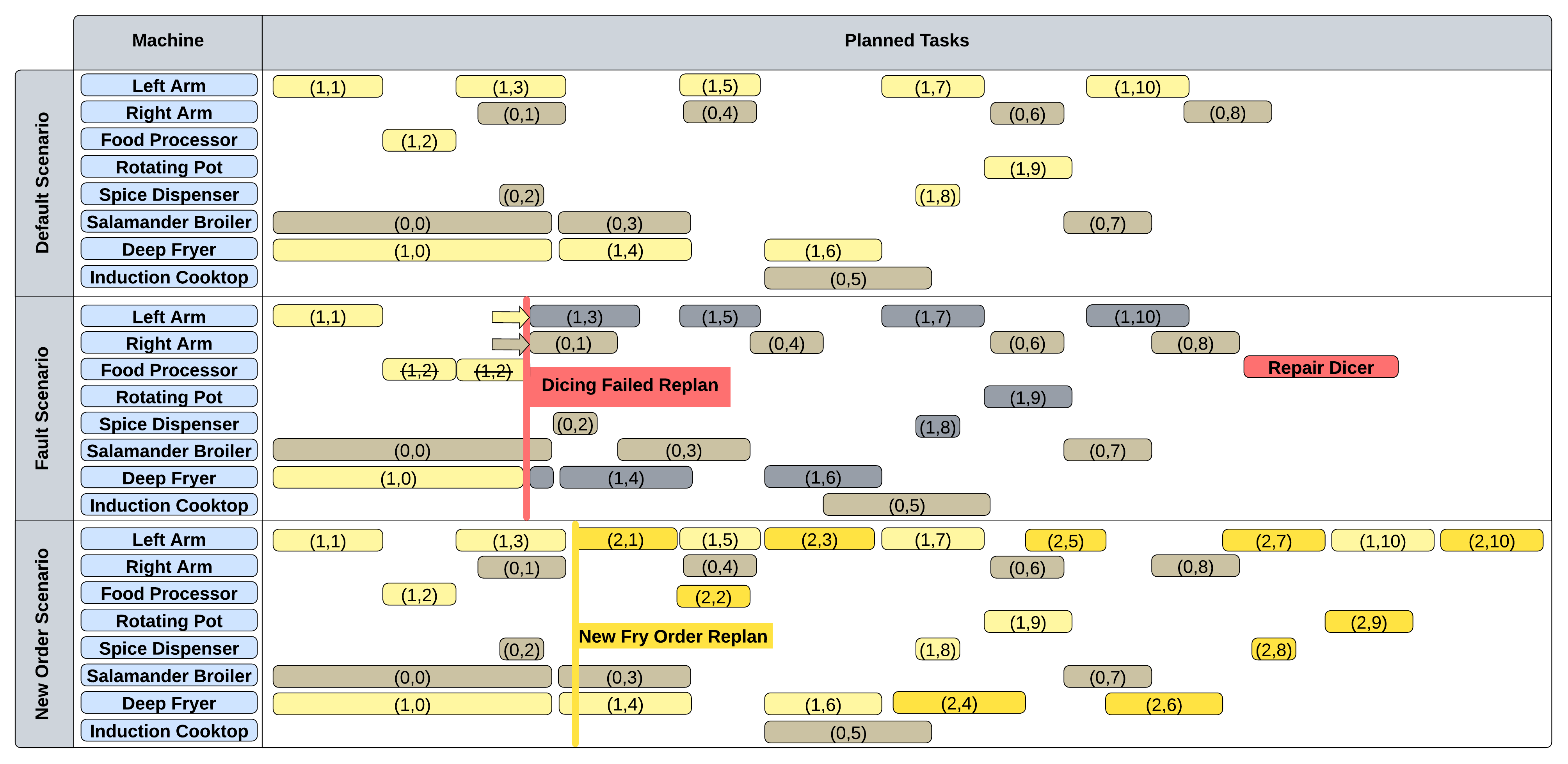}
\caption{The three Gantt charts above demonstrate the scheduling strategy when applied to 3 scenarios: a) the base example of one order of steak (brown) and one order of fries (yellow), b) a rescheduling of the same order after a failed dicing attempt with canceled tasks (grey), c) a rescheduling after a second order of fries (orange) is added to the queue part way through operation.}
\label{fig:Gantt}
\end{figure*}

New orders trigger rescheduling that accounts for unfinished and running tasks from previous orders. Priority maintains earlier order precedence while interweaving new tasks to minimize overall makespan. Fig.~\ref{fig:Gantt} illustrates this approach through three scenarios: base scheduling, failure recovery with canceled tasks, and mid-operation order addition for steak and fries preparation.

\section{Conclusion}\label{conclusion}

In this article, we have introduced the hardware and software design processes of an autonomous robotic cooking system utilizing cutting-edge technology. This system's development went beyond mere simulations and lab environments. The system was fully operational and publicly demonstrated at the Digital Innovation Festa 2023 in Seoul, Korea. To enhance its reliability, we simplified the system, which allowed us to successfully demonstrate the continuous cooking of steak frites numerous times throughout the convention. The system's modularity and strategic simplifications enabled this achievement, despite having only one day for preparation at the venue, underscoring the system's exceptional reliability and efficiency.

To provide quantitative validation of YORI's reliability, Table~\ref{tab:success_rate} presents the success rates across different deployment phases. The system successfully completed all conducted trials in these environments, from controlled laboratory settings to public demonstrations.

\begin{table}[t]
\centering
\caption{YORI Success Rate Across Deployment Phases}
\begin{tabular}{l c c}
\hline
\textbf{Phase} & \textbf{Location} & \textbf{Success Rate} \\
\hline
\hline
Integration Test & Research Lab & 30/30 \\
Post-Integration Rebuild & Sponsor HQ & 10/10 \\
Public Demo & Convention Center & 12/12 \\
\hline
\end{tabular}
\label{tab:success_rate}
\end{table}

\begin{table}[t]
\centering
\caption{Average Cooking Time with Parallel Execution}
\begin{tabular}{c c c}
\hline
\textbf{Number of} & \textbf{Total Time} & \textbf{Time per} \\
\textbf{Dishes} & \textbf{[min]} & \textbf{Dish [min]} \\
\hline
\hline
1 & 16.4& 16.4\\
2 & 30.1& 15.0\\
3 & 44.8& 14.9\\
\hline
\end{tabular}
\label{tab:cooking_time}
\end{table}

\section{Limitations}\label{limitations}
While YORI successfully demonstrates autonomous steak-frites cooking with high reliability and expandability for multi-dish preparation, its operation requires structured environmental conditions and lacks several capabilities necessary for universal deployment.

YORI's autonomy is currently limited to its environmental constraints. The system requires predetermined appliance positions and known ingredient locations. YORI cannot autonomously adapt to novel kitchen layouts, retrieve raw ingredients from unstructured storage, or navigate around dynamic obstacles.

The system's recovery capabilities currently remain limited. Despite retry logic and dynamic replanning for anticipated equipment failures, YORI lacks recovery mechanisms for unexpected events. Dropped ingredients, out of tolerance tool misalignment, or environmental changes require human intervention.

Furthermore, end-to-end automation remains incomplete. YORI assumes pre-portioned ingredients and does not address storage management, ingredient preparation, or post-cooking cleaning. These manual tasks prevent continuous autonomous operation in commercial settings.

These limitations reflect a deliberate design choice prioritizing developing core cooking capabilities with high reliability in controlled environments over broader autonomy with lower success rates. Advancing toward flexible autonomous cooking requires continued research addressing perception in unstructured environments, robust failure recovery, and the complete automation of ingredient storage through cleaning.

\section{Future Work}\label{future_work}
\subsubsection{Task \& Motion Planning}
Translating new recipes into executable tasks presents key challenges in developing methodologies that leverage robots' specific range of motion and customized tools and appliances designed for robotic manipulation, while considering both the available kitchen equipment and potential dual-arm coordination and multi-tasking capabilities. Additional challenges include preventing collisions between robotic arms and generating efficient trajectories through cluttered environments. Advancing these planning capabilities would enable the seamless integration of diverse recipes while maintaining operational safety and efficiency in robotic kitchen systems.

\subsubsection{Food Quality Control}
Autonomous cooking systems must verify ingredient freshness and determine cooking completion to ensure consistent quality. Visual inspection alone cannot fully assess the diverse set of data describing the state of an ingredient or dish. Machine learning models trained on multi-modal sensor data, particularly gas emissions during cooking, show promise for addressing these complexities. Such systems could enable real-time cooking adjustments based on unique ingredient characteristics, ensuring optimal quality and reducing waste.

\subsubsection{Cleaning \& Sanitation}
Autonomous cleaning capabilities are crucial for continuous operation in commercial settings. Future systems must handle cookware cleaning, surface sanitation, and waste disposal while adhering to food safety standards. Developing specialized end-effectors for scrubbing, rinsing, and sanitizing, combined with perception systems for cleanliness assessment, would enable fully autonomous kitchen operation without human intervention between cooking cycles.

\subsubsection{Learning-Based Approaches for Generalized Autonomy}
YORI's structured environment and deterministic control provide an ideal platform for collecting high-quality demonstration data. This data could train vision-based foundation models capable of high-level reasoning about recipe variations and robust motion planning and control. Such learning-based approaches could bridge the gap between YORI's structured automation and truly adaptive systems that approach human-level versatility in unstructured kitchens.

\section*{Acknowledgments}
This work was supported by Woowa Bros. Donghun Noh, Hyunwoo Nam, and Kyle Gillespie contributed equally to this project. We extend special thanks to Fadi Rafeedi for his assistance with tool manufacturing. We are also grateful to Naravit Vichathorn, Daehoon Kwon, and Alex Xu for their help in designing and fabricating the modular kitchen.

\bibliographystyle{IEEEtran}
\bibliography{yori}


\begin{IEEEbiographynophoto}{Donghun Noh}
Robotics and Mechanisms Laboratory, the Department of Mechanical and Aerospace Engineering, University of California, Los Angeles, CA 90095, USA. 
Email: donghun.noh@ucla.edu
\end{IEEEbiographynophoto}

\vskip -2\baselineskip plus -1fil

\begin{IEEEbiographynophoto}{Hyunwoo Nam}
Robotics and Mechanisms Laboratory, the Department of Mechanical and Aerospace Engineering, University of California, Los Angeles, CA 90095, USA. 
Email: harrynam@ucla.edu
\end{IEEEbiographynophoto}

\vskip -2\baselineskip plus -1fil

\begin{IEEEbiographynophoto}{Kyle Gillespie}
Robotics and Mechanisms Laboratory, the Department of Mechanical and Aerospace Engineering, University of California, Los Angeles, CA 90095, USA. 
Email: 	ktgillespie@ucla.edu
\end{IEEEbiographynophoto}

\vskip -2\baselineskip plus -1fil

\begin{IEEEbiographynophoto}{Yeting Liu}
Robotics and Mechanisms Laboratory, the Department of Mechanical and Aerospace Engineering, University of California, Los Angeles, CA 90095, USA. 
Email: liu1995@ucla.edu
\end{IEEEbiographynophoto}

\vskip -2\baselineskip plus -1fil

\begin{IEEEbiographynophoto}{Dennis Hong}
Robotics and Mechanisms Laboratory, the Department of Mechanical and Aerospace Engineering, University of California, Los Angeles, CA 90095, USA. 
Email: dennishong@ucla.edu
\end{IEEEbiographynophoto}




\vfill

\end{document}